\documentclass[sigconf]{acmart}

\usepackage{multirow}
\usepackage{booktabs}
\usepackage{tabularx}
\usepackage{threeparttable}
\usepackage{caption}
\usepackage{graphicx}
\usepackage{subfigure}
\usepackage{balance}

\AtBeginDocument{%
  }

\copyrightyear{2023}
\acmYear{2023}
\setcopyright{acmlicensed}
\acmConference[MAD '23]{2nd ACM International Workshop on Multimedia AI against Disinformation }{June 12, 2023}{Thessaloniki, Greece}
\acmBooktitle{2nd ACM International Workshop on Multimedia AI against Disinformation (MAD '23), June 12, 2023, Thessaloniki, Greece} \acmPrice{15.00}
\acmDOI{10.1145/3592572.3592846} \acmISBN{979-8-4007-0187-0/23/06}

\begin{document}

\title{Improving Synthetically Generated Image Detection in Cross-Concept Settings}

\author{Pantelis Dogoulis}

\orcid{1234-5678-9012}
\affiliation{%
  \institution{Information Technologies Institute, CERTH}
  \city{Thermi, Thessaloniki}
  \country{Greece}
}
\email{dogoulis@iti.gr}

\author{Giorgos Kordopatis-Zilos}
\affiliation{%
  \institution{Czech Technical University in Prague \\ Information Technologies Institute, CERTH}
  \city{}
  \country{}
}
\email{kordogeo@fel.cvut.cz}

\author{Ioannis Kompatsiaris}
\affiliation{%
  \institution{Information Technologies Institute, CERTH}
  \city{Thermi, Thessaloniki}
  \country{Greece}
}
\email{ikom@iti.gr}

\author{Symeon Papadopoulos}
\affiliation{%
  \institution{Information Technologies Institute, CERTH}
  \city{Thermi, Thessaloniki}
  \country{Greece}
}
\email{papadop@iti.gr}

\renewcommand{\shortauthors}{Pantelis Dogoulis et al.}

\begin{abstract}
New advancements for the detection of synthetic images are critical for fighting disinformation, as the capabilities of generative AI models continuously evolve and can lead to hyper-realistic synthetic imagery at unprecedented scale and speed. In this paper, we focus on the challenge of generalizing across different concept classes, e.g., when training a detector on human faces and testing on synthetic animal images -- highlighting the ineffectiveness of existing approaches that randomly sample generated images to train their models. By contrast, we propose an approach based on the premise that the robustness of the detector can be enhanced by training it on realistic synthetic images that are selected based on their quality scores according to a probabilistic quality estimation model. We demonstrate the effectiveness of the proposed approach by conducting experiments with generated images from two seminal architectures, StyleGAN2 and Latent Diffusion, and using three different concepts for each, so as to measure the cross-concept generalization ability. Our results show that our quality-based sampling method leads to higher detection performance for nearly all concepts, improving the overall effectiveness of the synthetic image detectors. \looseness=-1

\end{abstract}



\maketitle

\begin{figure}[t]
    \centering
    \includegraphics[width=0.87\linewidth]{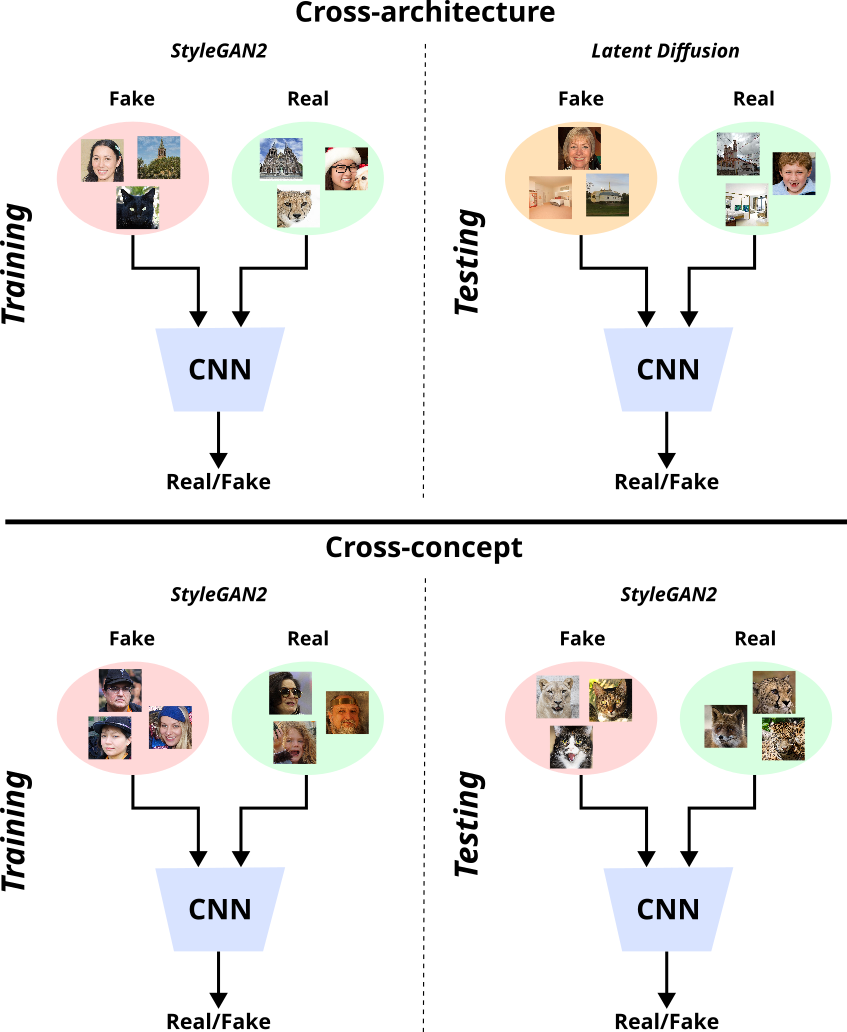}
    \vspace{-7pt}
    \caption{Overview of two generalization evaluation settings: \textit{Cross-architecture} (top), where training and testing are performed on images generated by different generative models, \textit{Cross-concept} (bottom), where training and testing are performed on images of different concepts generated by the same generative model.
    \vspace{-10pt}}
    \label{fig:fig_1}
\end{figure}

\section{Introduction}
Generative AI models constitute a rapidly growing area, and have already found their place in a wide range of applications \cite{ramesh2022hierarchical}. Advances have been driven in part by the development of new algorithms and techniques, such as Generative Adversarial Networks (GANs)~\cite{karras2018progressive, goodfellow2020generative, karras2019style,karras2020analyzing, karras2021alias} and recently Diffusion Models \cite{ho2020denoising,dhariwal2021diffusion, song2021scorebased,rombach2022high}, which have made it possible to generate increasingly realistic and high-quality synthetic data, reaching unprecedented levels of photorealism. At the same time, generative models can potentially be used by malicious users that spread disinformation across several social platforms. Also, a major concern is that recognizing whether an image is real or fake becomes increasingly challenging~\cite{farid2016photo}.

Distinguishing between generated and authentic images has attracted the interest of numerous researchers in the field of multimedia forensics~\cite{wang2022gan, gragnaniello2021gan}. The most widely used approach for detecting generated images involves training a neural network on a binary classification task (real vs fake) using a large corpus of labelled images. A key component in this development process is the application of a set of carefully selected image augmentations during the training phase, as demonstrated in prior work~\cite{wang2022gan, gragnaniello2021gan}. Our proposed approach relies on this supervised learning setup. 

Additionally, the generalization of the detectors are mostly studied with respect to the different generative architectures~\cite{wang2020cnn, gragnaniello2021gan}. The main goal is to create a detector that is effective in detecting fake images from different architectures.
However, herein we take a different avenue and define generalization as the ability to detect synthesized images that depict different concept classes, not necessarily generated from a different architecture (Figure \ref{fig:fig_1}). We refer to this capability as generalization in \emph{cross-concept settings}, which is often referred to as \emph{Domain Generalization} in literature~\cite{torralba2011unbiased}. We study the behavior of the detector in such a way that it could be trained in one concept class of real and fake images, e.g., \emph{human faces}, and can then be used to distinguish between real and fake images also in other concepts, e.g., \emph{animal faces}. We empirically find that using standard practices from the literature to train our detectors is not effective enough to generalize on cross-concept scenarios. \looseness=-1

In this paper, we tackle the cross-concept generalization challenge of synthetic image detectors by proposing a sampling strategy for the selection of generated images used for training. Prior work~\cite{wang2020cnn, gragnaniello2021gan} relies on generating a large number of images that are used for training. This is equivalent to random sampling, as no criterion for the selection of the generated images is applied. Instead, we assess the quality of the images based on a probabilistic method that provides a Quality Calculation (QC) score. Hence, we rank a large pool of generated images according to their scores and then select the top-k images in terms of quality to train our detectors. This is under the assumption that high  quality images will lead the network to focus less on the artifacts of the generative process and more on the characteristics that are invariant to the image content.
In that way, we should be able to build more robust detectors in the cross-concept scenario. We evaluate our method using fake images generated by StyleGAN2~\cite{karras2021alias} and the unconditional module of the recently introduced Latent Diffusion~\cite{rombach2022high}. We also evaluate on three concept classes for each generative architecture to evaluate cross-concept generalization. When training with the proposed sampling strategy, the performance is considerably improved compared to random selection. 

Our contribution can be summarized in the following:
\begin{itemize}
    \item We demonstrate the lack of generalization of state-of-the-art detectors in the cross-concept scenario.
    \item We propose a sampling strategy that considers  image quality scoring for sampling training data.
    \item We demonstrate improved performance using the proposed approach in the cross-concept settings of three concept classes for two generative architectures.
    \item We provide our code publicly available to facilitate future research on the field\footnote{\url{https://github.com/dogoulis/qc-sgid}}.
\end{itemize}

\section{Related Work}

Detecting synthetically generated images has become an increasingly important task due to the widespread use of numerous generative models to create fake images. In recent years, the research community has proposed various methods for detecting such images. In this section, we discuss the two main categories of methods that have been proposed in the literature: feature-based and frequency-based. Moreover, we analyze two recent works that also consider generated images based on diffusion models.

\subsection{Feature-based methods}

Feature-based methods for detecting synthetic images focus on extracting visual features from the images and using them to train a classifier. One of the most well-known methods in this category is the use of Convolutional Neural Networks (CNNs) trained on a large dataset of real and fake images~\cite{wang2020cnn}. In this work, Wang et al. show that a simple ResNet-50 classifier trained with a strong augmentation scheme can generalize to different generative architectures with high scores. Furthermore, \citet{gragnaniello2021gan} propose some modifications to the network architecture and pre-processing pipeline to improve the performance of the ResNet-50 classifier. The relation between augmentations during training and detection performance is also analyzed in~\cite{9897310}. Moreover, \citet{chai2020makes} propose using a patch-based CNN that focuses on local regions for improved performance. \citet{du2019towards} follow a similar local-based approach, training an autoencoder architecture that produces more generalizable results. Similarly, \citet{ju2022fusing} achieve good generalization by analyzing both local (patch-based feature selection) and global features (spatial information). An alternative approach is introduced in \cite{zhang2022improving}, where the authors construct a novel loss function under an unsupervised learning framework to boost the generalization of the detector. Specifically, they propose a contrastive loss function that encourages the network to learn discriminative features for real and fake images using a small number of unlabeled images from the target domain.  

\begin{figure*}[t]
    \centering
    \includegraphics[width=\textwidth]{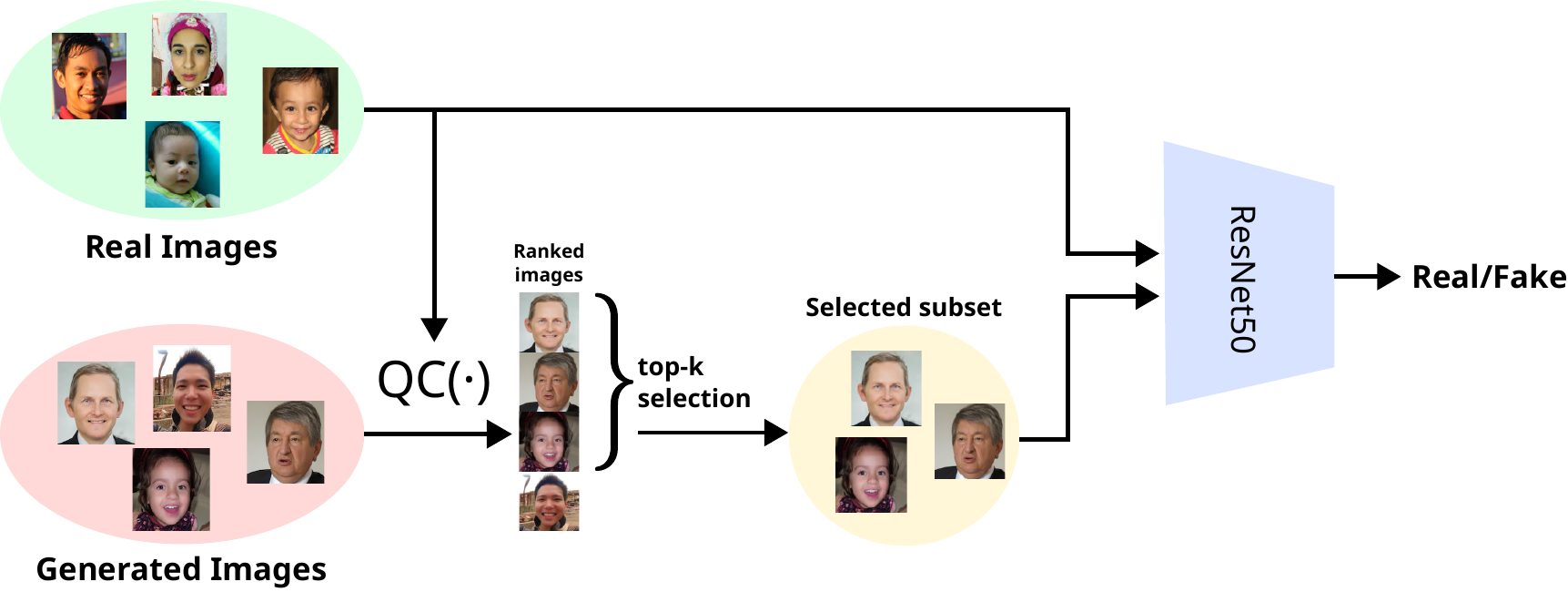}
    \caption{Overview of the proposed approach. The synthetically generated images (in the pink ellipsis) are evaluated based on a quality assessment model ($QC(\cdot)$) trained on real images (in the green ellipsis). The top-$k$ generated images are then selected and provided, along with real images, as input to a ResNet-50 model, which is trained to discriminate between real and fake images.\looseness=-1}
    \label{fig:scheme}
\end{figure*}

\subsection{Frequency-based methods}

Frequency-based methods for detecting synthetic images focus on analyzing the color and frequency characteristics of the images. Initial approaches establish the presence of artifacts that are both related to the generative model as well as its specific parameters \cite{marra2019gans, yu2019attributing}. One common approach is to extract co-occurrence matrices in both the spatial and frequency domains and feed them as input to CNN models~\cite{barni2020cnn, nataraj2019detecting}. Both of these works report good results with respect to the generalization of the detector across different generative architectures. \citet{zhang2019detecting} propose detecting artifacts in the spectrum space and use them as input to a ResNet34 network. Similarly, in \cite{frank2020leveraging}, the authors show that upsampling operations create artifacts in the frequency domain and optimize a simple classifier to detect synthetically generated images. A different approach is followed in \cite{liu2022detecting}, where the authors observe the existence of geometric grids in the magnitude and phase spectrograms of the generated images and use them for the detection of fake images.

\subsection{Detection of diffusion generated images}

Recently, the advent of diffusion models has stimulated the need for new methods to detect the corresponding synthetic images. \citet{corvi2022detection} conduct a comparative study of state-of-the-art detectors for GAN-generated images, evaluating their performance on the recently introduced diffusion-based generative methods. They show that current detectors cannot generalize to diffusion-based generative architectures. Similarly, \citet{sha2022fake} perform a comparative analysis of current state-of-the-art detectors and conclude that they are not effective in detecting images generated by recent GAN models. They also propose a multimodal detector that uses images and text prompts as input and show that this technique can improve detection performance. 

Overall, while there have been many proposed methods for detecting synthetic images, the task remains challenging, and new techniques are needed to keep up with the rapidly evolving landscape of generative models. However, none of these works address the problem of generalization across different concepts. 

\section{Proposed approach}
Our method can be split into two main parts. The first part is the sampling strategy, where the quality of the generated images is quantified and the best of them are selected to compose the training set. The second part is the training phase, where we use a CNN network and a pre-processing pipeline following the one proposed by two seminal works in the literature \cite{wang2020cnn, gragnaniello2021gan}. The use of quality quantification for composing the training set is arguably the most important part of our contribution. This builds on the intuitive assumption that if a network is trained on fake images of high perceptual quality, then it will focus on less obvious details and artifacts of the generative process. Hence, it will learn characteristics that are invariant to the content of the image and, as a result, boost the ability to detect fake images from different concepts.

\subsection{Quantifying quality in generated images}

There are different ways to assess the quality of AI-generated images automatically. The most well-known such metrics are Frechet Inception Distance (FID) \cite{heusel2017gans} and the Inception Score (IS) \cite{salimans2016improved}. The first measures the distance between the distributions of real and generated images in the feature space. The second measures the quality and diversity of generated images by looking at how well they are classified by a pre-trained Inception network~\cite{szegedy2015going}. This metric is generally more sensitive to changes in image quality than the FID. However, both methods measure the similarity between distributions of real and fake images, so they do not fit our needs. In this work, we use a metric that has been proposed in~\cite{gu2020giqa}, and we refer to it as Quality Calculation (QC). The intuition behind this approach is similar to the one behind the FID and IS. The quality of  generated images is quantified based on their similarity to a pre-defined real image distribution. First, a feature extractor $f(\cdot)$ is used to extract features from the real images. Then, the real image distribution is approximated by a Gaussian Mixture Model (GMM), which is a weighted sum of Gaussian distributions with parameters $\boldsymbol{\mu}^i$ and $\boldsymbol{\Sigma}^i$. Hence, the quality of the input image is calculated using the trained GMM. More formally, for an input image $I$, the quality score of the image is calculated based on:
\begin{equation}
    p(\textbf{x}|\lambda) = \sum_{i=1}^{M}\textbf{w}^i g(\textbf{x}|\boldsymbol{\mu}^i, \boldsymbol{\Sigma}^i)
\end{equation}
where $\boldsymbol{x} = f(I)$ and $\textbf{w}^i$ are the weights of the mixture model that satisfy the constraint $\sum_{i=1}^{M}\textbf{w}^i = 1$. The function $g(\cdot)$ represents the Gaussian density components of the model parameterized by the mean vector $\boldsymbol{\mu}^i$ and the covariance matrix $\boldsymbol{\Sigma}^i$. In a more general form, the parameters of this model can be denoted as $\boldsymbol{\lambda} = \{\boldsymbol{w}^i, \boldsymbol{\mu}^i, \boldsymbol{\Sigma}^i\}$. The estimation of $\lambda$ is given by the Expectation–Maximization (EM) algorithm \cite{dempster1977maximum}. Then, for each generated image $I_g$, its quality score $QC(I_g)$ is given by:
\begin{equation}
    QC(I_g) = p(f(I_g)|\boldsymbol{\lambda})
\end{equation}
The advantage of using the QC method when calculating the quality of the generated images is that it can produce a final ranking of images. Contrary to the FID score, which produces a single scalar that denotes the quality of a set of generated images, QC ranks the generated images from highest to lowest quality. 

\begin{table*}[t]
  \centering
  \begin{tabular}{ccccc}
    &  & \multicolumn{3}{c}{\textbf{test}} \\ \cmidrule{3-5}
    \textbf{training} & \textbf{sampling} & $\mathcal{H}$+$\mathcal{H}^G$ & $\mathcal{A}$+$\mathcal{A}^G$ & $\mathcal{C}$+$\mathcal{C}^G$\\
    \midrule
    \multirow{2}{*}{$\mathcal{H}$+$\mathcal{H}^G$} & random & \textbf{\color{gray}100.0$\pm$\color{gray}0.0} & 89.1$\pm$1.00 & 77.5$\pm$3.42 \\
    & QC (ours) & \textbf{\color{gray}100.0$\pm$\color{gray}0.0} & \textbf{94.8$\pm$1.22} & \textbf{83.5$\pm$0.94}  \\
    \midrule
    \multirow{2}{*}{$\mathcal{A}$+$\mathcal{A}^G$} & random & 61.8$\pm$1.36 & \textbf{\color{gray}100.0$\pm$\color{gray}0.0} & \textbf{89.1$\pm$0.71} \\
    & QC (ours) & \textbf{66.4$\pm$1.08} & \textbf{\color{gray}100.0$\pm$\color{gray}0.0} & 86.5$\pm$0.15 \\
    \midrule
    \multirow{2}{*}{$\mathcal{C}$+$\mathcal{C}^G$} & random & 53.7$\pm$0.22 & 58.6$\pm$3.51 & \textbf{\color{gray}100.0$\pm$\color{gray}0.0} \\
    & QC (ours) & \textbf{59.4$\pm$0.19} & \textbf{67.1$\pm$0.16} & \textbf{\color{gray}100.0$\pm$\color{gray}0.0} \\
    \bottomrule
    \\
    \multicolumn{5}{c}{(a) StyleGAN2} \\
  \end{tabular} \hspace{0.5cm}
  \begin{tabular}{ccccc}
    &  & \multicolumn{3}{c}{\textbf{test}} \\ \cmidrule{3-5}
    \textbf{training} & \textbf{sampling} & $\mathcal{H}$+$\mathcal{H}^D$ & $\mathcal{B}$+$\mathcal{B}^D$ & $\mathcal{C}$+$\mathcal{C}^D$\\
    \midrule
    \multirow{2}{*}{$\mathcal{H}$+$\mathcal{H}^D$} & random & \textbf{\color{gray}100.0$\pm$\color{gray}0.0} & 64.4$\pm$2.31 & 66.7$\pm$0.88 \\
    & QC (ours) & \textbf{\color{gray}100.0$\pm$\color{gray}0.0} & \textbf{74.1$\pm$0.68} & \textbf{77.5$\pm$0.03} \\
    \midrule
    \multirow{2}{*}{$\mathcal{B}$+$\mathcal{B}^D$} & random & 52.1$\pm$1.47 & \color{gray}96.3$\pm$\color{gray}0.79 & \textbf{99.7$\pm$0.03} \\
    & QC (ours) & \textbf{56.2$\pm$5.13} & \textbf{\color{gray}99.6$\pm$\color{gray}0.02} & 99.4$\pm$0.02 \\
    \midrule
    \multirow{2}{*}{$\mathcal{C}$+$\mathcal{C}^D$} & random & 54.2$\pm$4.08 & 96.3$\pm$1.19 & \textbf{\color{gray}100.0$\pm\color{gray}$0.0} \\
    & QC (ours) & \textbf{58.5$\pm$1.52} & \textbf{98.9$\pm$0.58} & \textbf{\color{gray}100.0$\pm$\color{gray}0.0} \\
    \bottomrule
    \\
    \multicolumn{5}{c}{(b) Latent Diffusion} 
  \end{tabular}
  \caption{AUC of detection model trained on randomly selected samples or based on the proposed QC score for each concept and generative model. Mean and standard deviation of three training sessions with different seeds are reported. \textbf{Bold} indicates the best performance between the proposed and the random baseline. {\color{gray} Gray} colour indicates intra-concept evaluation.
\vspace{-15pt}
}
\label{combined-table}
\end{table*}

\subsection{Synthetically generated image detection}
\label{sec:method}

Figure \ref{fig:scheme} displays an overview of the proposed approach. First, we train the QC models~\cite{gu2020giqa} for each of the different concepts based on the corresponding datasets of real images. Then, we select the top-$k$ generated images with the highest quality and the same amount of real images in order to form the training dataset. The next step is to train a classifier network to distinguish between real and fake images. The classifier network and the training process are based on the methodology proposed in~\cite{wang2020cnn}. A ResNet-50~\cite{he2016deep} model is initialized with the weights of ImageNet~\cite{5206848} and trained using the dataset compiled in the previous step. Additionally, during training, we apply a set of augmentations including several geometric augmentations (i.e., random crop and resize), Gaussian blurring, and JPEG compression. The network minimizes the binary cross-entropy loss. \looseness=-1

\section{Evaluation setup}

In this section, we discuss the evaluation pipeline for our experiments. Specifically, we present the datasets, evaluation methodology, and implementation details used.

\subsection{Datasets}
We obtain real images from publicly available datasets, including FFHQ~\cite{karras2019style}, AFHQ~\cite{choi2018stargan}, and LSUN~\cite{yu2015lsun}. The FFHQ dataset contains images of human faces, the AFHQ dataset contains images of dogs, cats, and a general class of wildlife, while the LSUN dataset contains nearly one million images of 10 scene categories and twenty object classes. Then, we employ pretrained diffusion and GAN models to generate artificially generated images for different classes to evaluate the cross-concept scenario. More specifically, we use the StyleGAN2~\cite{karras2020analyzing} model to generate images from pretrained networks in FFHQ, AFHQ and LSUN-churches, while we also use the Latent Diffusion~\cite{rombach2022high} model to generate images from pretrained networks in FFHQ, LSUN-bedrooms, and LSUN-churches. We denote each dataset as $\mathcal{X} + \mathcal{X}^A$, where $\mathcal{X}$ is the set of real images of a specific concept class and $\mathcal{X}^A$ is the corresponding set of fake images generated with architecture $A$, which takes values either $G$ for StyleGAN2 or $D$ for Latent Diffusion. Regarding the different concepts, we use $\mathcal{H}$ for human, $\mathcal{A}$ for animal, $\mathcal{C}$ for church, and $\mathcal{B}$ for bedroom datasets.

\subsection{Evaluation methodology and metric}

We assess the detection performance of our model on each concept class by utilizing the AUC metric. This particular scoring method does not rely on any predetermined threshold, making it suitable for assessing the robustness and generalization of the detector models. 
For each concept, we perform three training sessions and report the mean AUC and its standard deviation.  
For comparison, we train the ResNet-50 classifier following the same training process but with random sampling for fake images instead, i.e., without using the proposed sampling strategy. 

\subsection{Implementation details}

For our sampling strategy, we initially generate 20K total images for each concept and architecture, and then select the top 10K images based on their QC score. The test set consists of 2K images for each concept and it is randomly selected before the QC selection. We sampled the same number of real images in order to form the training and test datasets. The GMM model that was implemented, used the Inception model~\cite{szegedy2015going} as a feature extractor. Moreover, we used 50 Gaussian components and a batch size of 50 instances.

For preprocessing, during training, we implement the augmentations from Section \ref{sec:method} with an output size of $224 \times 224$ pixels. While during testing, we only resize images to $224 \times 224$.

For our detector, we use a ResNet-50~\cite{he2016deep} pretrained on ImageNet~\cite{5206848}. The model is trained using the AdamW optimizer \cite{loshchilov2018decoupled}. The learning rate was equal to $10^{-3}$, and a step scheduler with 5 epochs was used. Weight decay is also applied with a factor of $5\cdot 10^{-5}$. Additionally, a drop path~\cite{huang2016deep} rate of $0.1$ is employed to prevent overfitting, which randomly drops entire paths (i.e., sequences of layers) in the model during training. All training and evaluation processes were carried out on a server with one NVIDIA GeForce RTX 3060 GPU.

\section{Experiments}
\begin{table*}[t]
  \centering
  \scalebox{0.92}{
  \begin{tabular}{cccccc}
    &  & \multicolumn{4}{c}{\textbf{quality quartile $\downarrow$}} \\ \cmidrule{3-6}
    \textbf{training} & \textbf{test set} & 0-25\% & 25-50\% & 50-75\% & 75-100\%\\
    \midrule
    \multirow{2}{*}{$\mathcal{H}$+$\mathcal{H}^G$} & $\mathcal{A}$+$\mathcal{A}^G$ & 93.4$\pm$0.98 & 92.8$\pm$1.32 & 93.2$\pm$1.17 & 89.8$\pm$0.89\\
     & $\mathcal{C}$+$\mathcal{C}^G$ & 83.0$\pm$1.07 & 79.2$\pm$1.03 & 80.9$\pm$0.83 & 77.8$\pm$0.92\\  \midrule
    \multirow{2}{*}{$\mathcal{A}$+$\mathcal{A}^G$} & $\mathcal{H}$+$\mathcal{H}^G$ & 66.8$\pm$1.02 & 66.2$\pm$1.15 & 63.9$\pm$0.97 & 62.7$\pm$0.98\\
    & $\mathcal{C}$+$\mathcal{C}^G$ & 84.1$\pm$0.32 & 84.9$\pm$0.17 & 90.3$\pm$0.11 & 94.0$\pm$0.13 \\  \midrule
    \multirow{2}{*}{$\mathcal{C}$+$\mathcal{C}^G$} & $\mathcal{H}$+$\mathcal{H}^G$ & 56.1$\pm$0.19 & 59.9$\pm$0.18 & 59.5$\pm$0.21 & 58.8$\pm$0.17\\
    & $\mathcal{A}$+$\mathcal{A}^G$ & 62.2$\pm$0.13 & 64.1$\pm$0.17 & 67.1 $\pm$0.14 & 69.8$\pm$0.24 \\
    \bottomrule
    \\
    \multicolumn{6}{c}{(A) StyleGAN2} 
  \end{tabular} \hspace{0.3cm}
  \begin{tabular}{cccccc}
    &  & \multicolumn{4}{c}{\textbf{quality quartile $\downarrow$}} \\ \cmidrule{3-6}
    \textbf{training} & \textbf{test set} & 0-25\% & 25-50\% & 50-75\% & 75-100\%\\
    \midrule
    \multirow{2}{*}{$\mathcal{H}$+$\mathcal{H}^D$} & $\mathcal{B}$+$\mathcal{B}^D$ & 70.1$\pm$0.72 & 73.6$\pm$0.67 & 74.8$\pm$0.68 & 72.5$\pm$0.69\\
     & $\mathcal{C}$+$\mathcal{C}^D$ & 74.1$\pm$0.05 & 73.2$\pm$0.04 & 78.6$\pm$0.07 & 77.4$\pm$0.02\\  \midrule
    \multirow{2}{*}{$\mathcal{B}$+$\mathcal{B}^D$} & $\mathcal{H}$+$\mathcal{H}^D$ & 52.1$\pm$5.42 & 56.9$\pm$5.34 & 55.3$\pm$5.22 & 57.7$\pm$4.88\\
    & $\mathcal{C}$+$\mathcal{C}^D$ & 99.2$\pm$0.02 & 98.3$\pm$0.07 & 99.4$\pm$0.04 & 99.4$\pm$0.02 \\  \midrule
    \multirow{2}{*}{$\mathcal{C}$+$\mathcal{C}^D$} & $\mathcal{H}$+$\mathcal{H}^D$ & 55.7$\pm$1.55 & 56.4$\pm$1.52 & 57.8$\pm$1.48 & 57.3$\pm$1.42\\
    & $\mathcal{B}$+$\mathcal{B}^D$ & 98.2$\pm$0.64 & 98.4$\pm$0.72 & 98.4$\pm$0.58 & 99.3$\pm$0.71 \\
    \bottomrule
    \\
    \multicolumn{6}{c}{(b) Latent Diffusion} 
  \end{tabular}}
  \caption{AUC of detection model for test subsets grouped based on their QC score for the two generative models. Mean and standard deviation of three training sessions with different seeds are reported.
  \vspace{-5pt}}
\label{tab:test_giqa}
\end{table*}

In this section, we present the experimental results of the proposed method and demonstrate the generalization improvements of the models when trained with it. Our primary interest lies in the cross-concept scenario, where a network is trained on instances of a specific concept class and evaluated on unseen instances from a different concept class. This evaluation serves as a measure of the model's ability to generalize beyond its training domain and is a critical aspect of assessing the effectiveness of the proposed method in terms of cross-concept generalization.

\subsection{Cross-concept evaluation}

Table \ref{combined-table} presents the AUC scores of the detectors trained with random sampling and the proposed method, for generated images based on StyleGAN2 and Latent Diffusion. To ensure the validity and reliability of our findings, we run three training sessions for each concept and report the mean and standard deviation of AUC.

We first analyse the results of our intra-class experiments, where we train and evaluate a detector in the same concept class. The corresponding runs are coloured in gray in Table \ref{combined-table}. Our findings indicate that there is no significant difference in the performance of the detector when using images of higher quality, due to the perfect performance in almost all cases (with the exception of the Bedroom concept in Latent Diffusion). This result is consistent with previous studies that have shown that a robust augmentation scheme is sufficient to train very accurate detectors within the same domain \cite{wang2020cnn,gragnaniello2021gan}. In our study, we used the same augmentation scheme for both randomly selected and quality-based selected subsets, which explains the lack of notable difference in the detector's performance between the two subsets.

Next, we discuss the results of our proposed methodology in the case where a detector is trained on generated and real images from a concept class and is evaluated on images of a different class. Regarding the experiments on StyleGAN2 sets, it is evident that our proposed method clearly improves the robustness of the detector in almost all cases except in the case when training on Animals ($\mathcal{A}$+$\mathcal{A}^G$) and testing on Churches ($\mathcal{C}$+$\mathcal{C}^G$). In all other cases, our proposed approach outperforms the baseline by more than 5\% in terms of mean AUC, reaching up to 8.5\%. Furthermore, the standard deviation of the runs with the proposed QC sampling strategy is relatively low, being less than 1.5\% in all cases, and often significantly lower compared with the runs where random sampling was used. This implies that using our strategy leads to more consistent and reliable models. Similar conclusions can be drawn from the experiments on Latent Diffusion sets. Specifically, our method surpasses the baseline in all but one case, i.e., training on Bedrooms ($\mathcal{B}$+$\mathcal{B}^D$) and testing on Churches ($\mathcal{C}$+$\mathcal{C}^D$), where the difference is marginal. The AUC score is improved by almost 10\% in some cases. Also, the standard deviation is generally low, except for one case. \looseness=-1

In summary, our experimental comparison provides compelling evidence of the effectiveness of the proposed method in improving the generalization performance across different concepts.

\subsection{Results when test image quality varies}

We also evaluate our model in the case of selecting different test set composition in terms of image quality. Specifically, the aim is to evaluate whether images of better quality are more difficult to detect or not. Hence, we split the test sets into four quartiles denoted as 0-25\% (lowest quality), 25-50\%, 50-75\%, and 75-100\% (highest quality) based on their QC score. We observe that in several cases of StyleGAN2 generated images, the detector achieves better AUC scores for low- or medium-quality images. This means that a test set would benefit from an automatic selection step where higher quality images are retained, since it would be more challenging. 
Instead, in the case of images generated using the Latent Diffusion model, there is no apparent association between the test set composition in terms of image quality and the detection performance. \looseness=-1

\begin{figure*}[t]
    \centering
    \subfigure[StyleGAN2]{
    \begin{tabular}{ccc}
        Human ($\mathcal{H}^G$) & Animal ($\mathcal{A}^G$) & Church ($\mathcal{C}^G$) \\
        \includegraphics[width=0.25\linewidth]{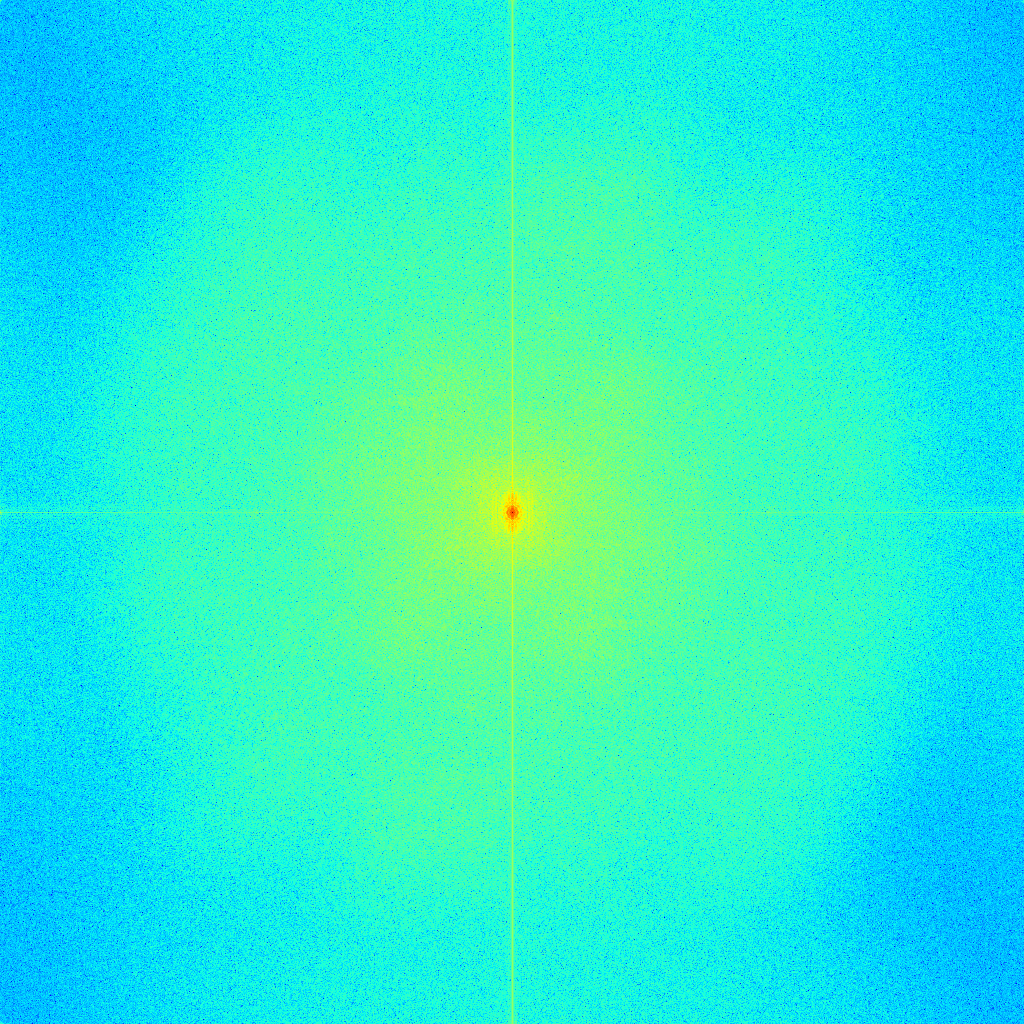}\label{fig:4} & 
        \includegraphics[width=0.25\linewidth]{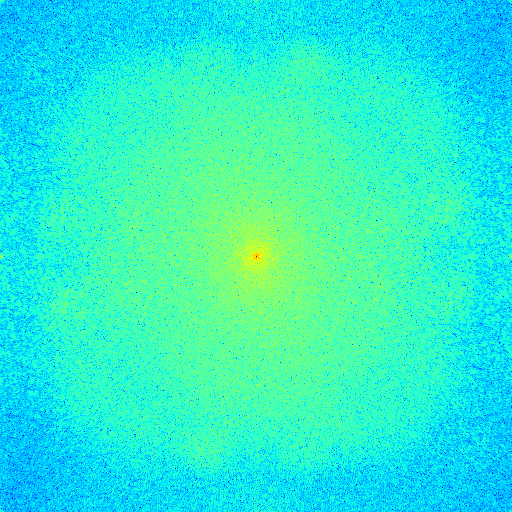}\label{fig:5} & 
        \includegraphics[width=0.25\linewidth]{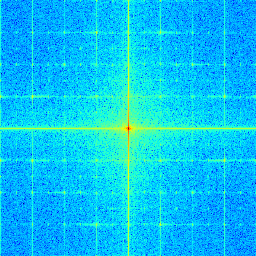}\label{fig:6} 
    \end{tabular}}
    \subfigure[Latent Diffusion]{
    \begin{tabular}{ccc}
        Human ($\mathcal{H}^D$) & Bedroom ($\mathcal{B}^D$) & Church ($\mathcal{C}^D$) \\
        \includegraphics[width=0.25\linewidth]{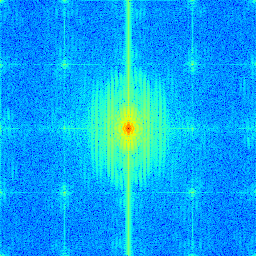}\label{fig:1} & 
        \includegraphics[width=0.25\linewidth]{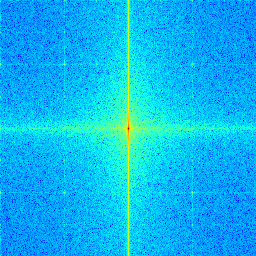}\label{fig:2} & 
        \includegraphics[width=0.25\linewidth]{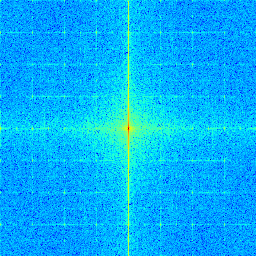}\label{fig:3} 
    \end{tabular}}
    \vspace{-10pt}
    \caption{Visualization of the magnitude spectrograms for the different classes of the two generative architectures using the denoising network proposed in \cite{zhang2017beyond}.
    }
    \label{fig:artifact-analysis}
\end{figure*}

\subsection{Results when the detector is trained with many classes and architectures}

\begin{table}[t]
  \centering
  \begin{tabular}{cccc}
    & \multicolumn{3}{c}{\textbf{test}} \\ \cmidrule{2-4}
    \textbf{training} & $\mathcal{G}$ & $\mathcal{D}$ & $\mathcal{G+D}$ \\
    \midrule
    $\mathcal{G}$ & 99.9$\pm$0.00 & 60.5$\pm$1.53 & 77.3$\pm$0.88 \\
    $\mathcal{D}$ & 59.2$\pm$0.29 & 99.9$\pm$0.00 & 74.6$\pm$1.07 \\
    $\mathcal{G+D}$ & 99.9$\pm$0.00 & 99.9$\pm$0.01 & 99.9$\pm$0.00 \\
    \bottomrule
  \end{tabular}
  \caption{AUC of detection model when trained on all the different classes and architectures. Mean and standard deviation of three sessions with different seeds are reported.
  \vspace{-15pt}}
  \label{table:agg-res}
\end{table}

In this section, we evaluate the performance of the model when trained based on QC sampling in three different datasets:

\begin{itemize}
    \item $\mathcal{G}$: Dataset consists of images from all concepts, generated by StyleGAN2, and the corresponding sets of real images.
    \item $\mathcal{D}$: Dataset consists of images from all concepts, generated by Latent Diffusion, and the corresponding sets of real images.
    \item $\mathcal{G+D}$: Combination of the other two datasets.
\end{itemize}

The findings are presented in Table \ref{table:agg-res}. Our analysis reveals that the performance of the detector significantly improves when it is trained on a diverse set of generated images, including those produced by StyleGAN2 and Latent Diffusion models, and evaluated on a range of test sets. Conversely, we observe that detectors trained solely on images generated by a single architecture demonstrate limited generalization ability when tested on images generated by other architectures \cite{sha2022fake, corvi2022detection}. Furthermore, our investigation indicates that the semantic content of each dataset is a crucial factor affecting the detection of generated images, and the detector's efficacy is significantly enhanced with the exploitation of a greater variety of classes during training.

\begin{figure*}[ht!]
    \centering
    \subfigure[StyleGAN2]{
    \includegraphics[width=0.45\linewidth]{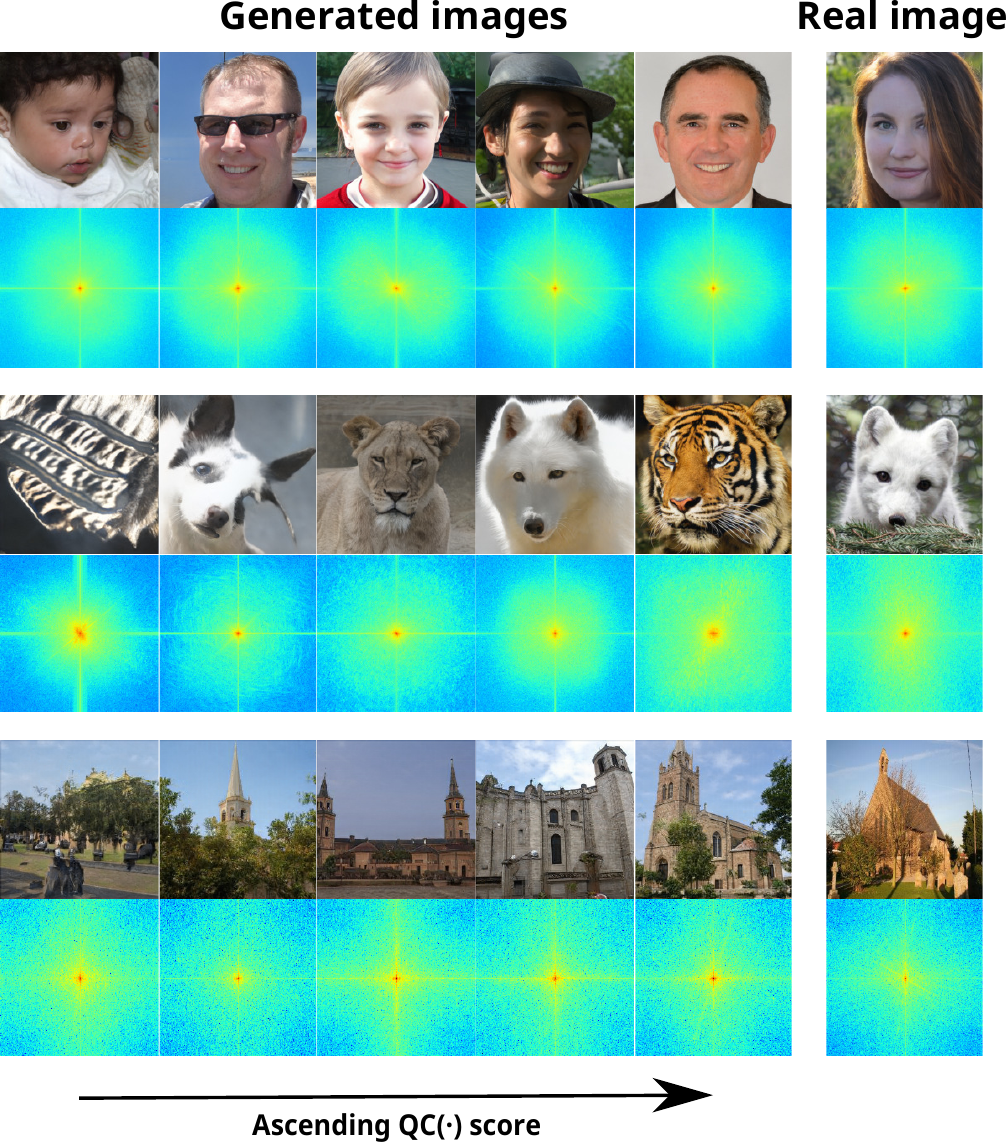} }
    \hspace{0.5cm}
    \subfigure[Latent Diffusion]{
    \includegraphics[width=0.45\linewidth]{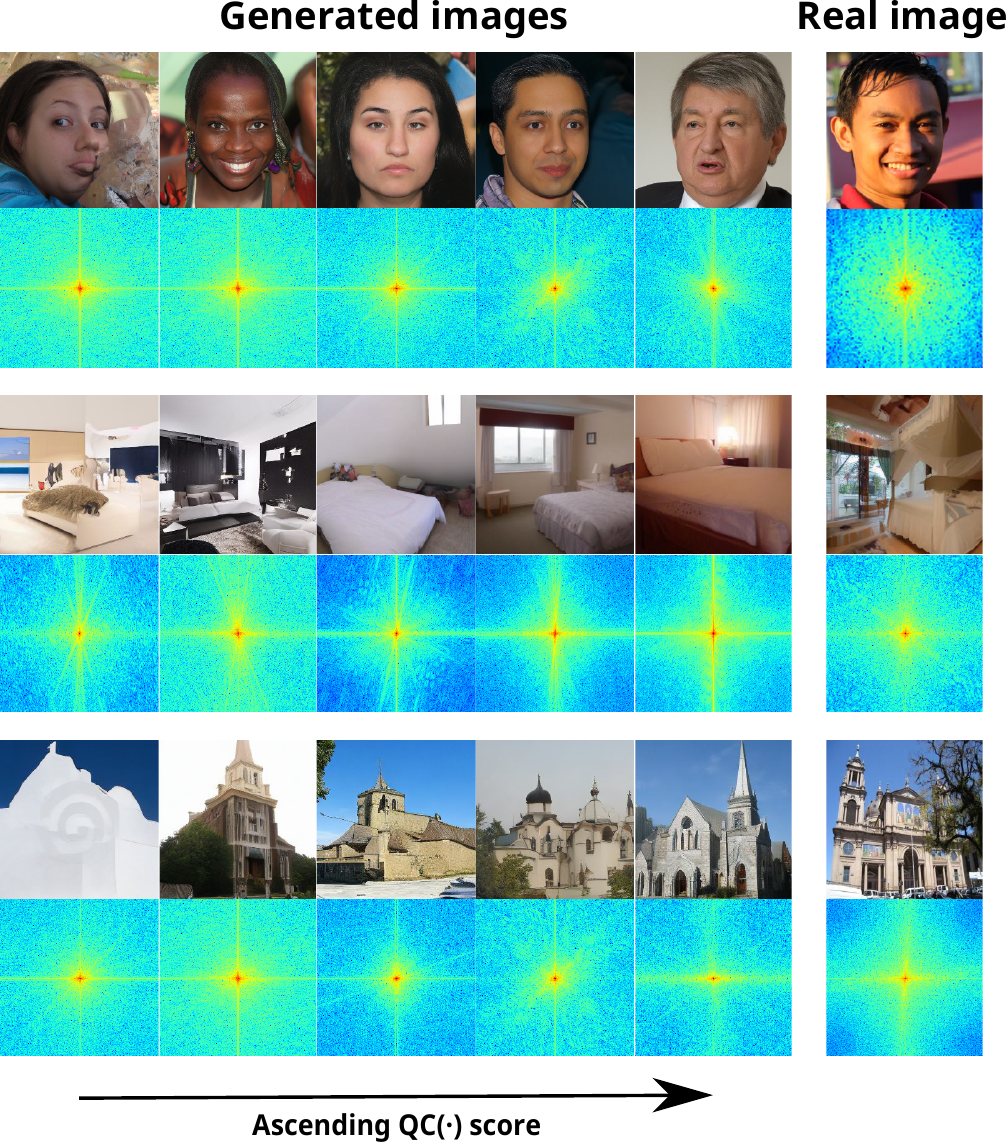} }
    \vspace{-10pt}
    \caption{Sampled image and magnitude spectrogram of the top 100\%, 75\%, 50\%, 25\%, 1\% generated images ranked based on their QC score and a real image from the same concept with its magnitude spectrogram.
    \vspace{-5pt}
    } 
    \label{fig:rank-giqa}
\end{figure*}

\subsection{Qualitative artifact analysis}

In this section, we visualize the artifacts that are calculated using the Fast Fourier Transform for different generated classes of the same architecture. Following \cite{corvi2022detection}, we first randomly select 1000 generated images per concent and model architecture. Then using the denoising function proposed in \cite{zhang2017beyond}, we transform each image $X$ to its denoised version $f(X)$ and then compute the residual $R(X) = X - f(X)$. Next, we average the residuals and apply a 2D Fast Fourier Transform in order to obtain the magnitude and phase spectrograms. The magnitude spectrograms are a good indicator for the analysis of the artifacts introduced by a generative model.

As seen in Figure \ref{fig:artifact-analysis}, differences exist between the generated images of each concept. It is evident that generated images from models pretrained on datasets with similar structures present more common characteristics. For instance, FFHQ and AFHQ, which are used for Humans $\mathcal{H}^G$ and Animals $\mathcal{A}^G$, respectively, are high-quality and high-resolution datasets consisting of images with centered and aligned faces. This produces similar cloudy magnitude spectrograms for these two cases. Similar spectrograms also appear in the case of the Bedrooms $\mathcal{B}^D$ and Churches $\mathcal{C}^D$ generated from the models that were trained on the LSUN-Bedroom and LSUN-Church datasets, which contain images of significantly lower resolution and quality compared to FFHQ and AFHQ. Nevertheless, all images are preprocessed in the same way in our experiments; hence, we expect such differences to be mitigated during our evaluations. This observation reinforces the main assumption of this work, which is that images produced by generative architectures are class-dependent. This means that the classification of real and fake images is affected by the semantic image content \cite{torralba2011unbiased,wang2020cnn}.

Moreover, Figure \ref{fig:rank-giqa} illustrates five generated images for each generation model and concept as well as one real image from the same concept. Specifically, we rank the generated images from each concept, and then we select the images ranked on 1\%, 25\%, 50\%, 75\%, and 100\% quartiles of the total image quality ranking. This means that the image ranked in the 1\% is the image with the best quality, the image ranked in the 100\% quartile has the worst quality, while the images located within the 25\%, 50\%, and 75\% quartiles denote images of progressively lower quality as their respective quartile rankings increase. It is noticeable that moving from the worst to the best image ranked by the QC score, the perceptual quality is significantly better. Furthermore, we provide the corresponding magnitude spectrograms of each image in the second row of the figure. Interestingly, we observe that the image with the best quality has a very similar spectrogram to the spectrogram of the real image, while as we move to the image with the worst quality, the spectrogram appears to be noisier. This finding supports our main intuition that training with higher-quality images can facilitate the learning of subtle and less distinguishable artifacts by the network.

\section{Conclusion}

In this work, we address the challenge of generalization for synthetic image detection in the cross-concept scenario. We observe that even state-of-the-art detectors that use strong augmentation schemes to improve their robustness to different generative architectures still lack the ability to generalize to unseen concept classes. To address this, we propose a novel approach that uses a probabilistic method to quantify the quality of generated images and shows that training detectors with higher-quality images can significantly improve their generalization ability. We evaluate our method using fake images generated by two state-of-the-art generative models, StyleGAN2 and Latent Diffusion, and demonstrate that our approach improves the detection performance when training with randomly selected generated images. Also, we enhance the basic intuition by visualising the magnitude spectrograms of quality-based selected generated images. Although our results are promising, there is still considerable room for improvement in the development of detectors that can generalize better across different concept classes and generative models. To this end, we plan to carry out more extensive experiments involving more concepts and generative models, and investigating the potential of frequency-based methods and ensembles of models to further increase the robustness and accuracy of the detectors.

\section{Acknowledgments}

This work was partially funded by the projects ``vera.ai: VERification Assisted by Artificial Intelligence'' (GA no. 101070093) and ``AI4Media: A European Excellence Centre for Media, Society and Democracy'' (GA no. 951911).

\balance
\bibliographystyle{ACM-Reference-Format}
\bibliography{sample-base.bib}

\end{document}